\documentclass[runningheads]{llncs}
\usepackage[T1]{fontenc}
\usepackage{graphicx,verbatim}
\usepackage{hyperref}
\usepackage{color}

\usepackage{array}
\usepackage{booktabs}
\usepackage{graphicx}
\usepackage{makecell}
\usepackage{multirow}
\usepackage{rotating}

\newcommand{\grouplabel}[2]{%
  \multirow{#1}{*}{\parbox[c][\dimexpr #1\ht\strutbox + #1\dp\strutbox\relax][c]{\linewidth}{%
    \centering\rotatebox{90}{\scriptsize\scshape #2}}}%
}

\usepackage{amsmath}

\begin{document}
\title{A Multimodal 3D Foundation Model for Light Sheet Fluorescence Microscopy Enables Few-Shot Segmentation, Classification, and Deblurring\thanks{Accepted at MICCAI 2026.}}
\titlerunning{A Multimodal 3D Foundation Model for Light Sheet Microscopy}
%
\author{Adina Scheinfeld\inst{1,2} \and Haotan Zhang\inst{3,4} \and Shang Mu\inst{3} \and Rudolf L. M. van Herten\inst{2,5} \and Lucas Stoffl\inst{2,5} \and Ali Ertürk\inst{6,7} \and Zhuhao Wu\inst{3} \and Johannes C. Paetzold\inst{2,5}}
\authorrunning{A. Scheinfeld et al.}
%
\institute{Tri-Institutional Program in Computational Biology \& Medicine, Weill Cornell Medicine, New York, NY, USA \\ \email{ads4015@med.cornell.edu}  
\and Department of Radiology, Weill Cornell Medicine, New York, NY, USA 
\and Helen and Robert Appel Alzheimers Disease Research Institute, Feil Family Brain and Mind Research Institute, Weill Cornell Medicine, New York, NY, USA 
\and Graduate Program in Physiology, Biophysics and Systems Biology, Weill Cornell Medicine, New York, NY, USA 
\and Cornell Tech, New York, NY, USA 
\and Institute for Intelligent Biotechnologies (iBIO), Helmholtz Center Munich, Neuherberg, Germany 
\and Institute for Stroke and Dementia Research, Klinikum der Universität München, Ludwig-Maximilians University Munich, Munich, Germany 
} 

  
\maketitle              

\begin{abstract}
Light sheet fluorescence microscopy (LSM) enables\break high-resolution, three-dimensional (3D) imaging of biological specimens, providing rich volumetric data for studying cellular organization, pathology, and vascular networks. However, the size, dimensionality, and annotation burden of LSM data make supervised deep learning approaches costly and difficult to scale. Additionally, despite the abundance of unannotated LSM volumes, foundation models for this modality remain underexplored due to computational challenges and the complexity of volumetric representation learning. In this work, we introduce a 3D foundation model for LSM data, pretrained on a large curated collection of 3D images spanning multiple organisms, stains, and imaging protocols. We learn transferable volumetric representations by jointly optimizing for masked reconstruction and image-text alignment. The pretrained backbone drastically reduces the annotation burden, enabling efficient, few-shot adaptation for varied downstream tasks. We evaluate this approach on downstream segmentation, classification, and deblurring. Our results demonstrate consistent improvements over baselines, (1) when measured using standard evaluation metrics and (2) when rigorously assessed by domain experts. This highlights the potential of foundation model pretraining to reduce annotation requirements while improving performance across diverse LSM analysis tasks. Pretrained model weights and code for pretraining and finetuning are publicly available: \url{https://github.com/AdinaScheinfeld/lsm_fm_public_repo.git}.
\keywords{Foundation Models \and LSM \and SSL \and Multimodal Learning.}
\end{abstract}

\section{Introduction}


Light sheet fluorescence microscopy has become a valuable imaging tool, enabling the rapid acquisition of high-resolution 3D images at whole-organ \cite{ref_lsm,ref_vr_cells_kaltenecker,ref_intact_organs_zhao} and whole-organism scale \cite{ref_wholebody_igg_mai}. The size and quality of these images enable analysis of cellular structures, pathological markers, and complex networks, yet also present challenges for downstream analysis due to the time and expertise required. And while deep learning methods have shown promise in accelerating such analyses, their success typically depends on large quantities of annotated training data \cite{ref_3d_lack,ref_cellpose}. For LSM, generating annotations is difficult as segmentation requires voxel-level labels, image classification requires expert knowledge of stains and structures, and image restoration tasks require paired degraded and high-quality volumes. Therefore, most existing models are trained for narrow tasks (e.g. segmentation only \cite{ref_cellseg,ref_microsam,ref_cellpose}) or for specific structures (e.g. nuclei or vessels only \cite{ref_cellseg,ref_cellpose,ref_vessap}) and fail to generalize across diverse staining protocols and biological contexts.

Foundation models offer a compelling alternative by learning transferable representations from large-scale unlabeled data through self-supervised objectives \cite{ref_fm_bommasani}. Such models have achieved success in domains including natural images \cite{ref_natimages_kirillov,ref_natimages_wang}, medical imaging \cite{ref_medimaging_ma,ref_medimaging_zhao,ref_medimaging_zhou}, genomics \cite{ref_genomics_cui}, language \cite{ref_language_grattafiori}, and agentic clinical reasoning \cite{ref_veritas}. In microscopy however, foundation model development remains limited, particularly for volumetric light sheet data. Owing to its high signal-to-noise ratio and the abundance of unannotated volumetric data routinely generated in biological studies, LSM presents unique, underexplored opportunities for unsupervised learning. 

\begin{figure} [t]
\centering
    \includegraphics[width=0.93\textwidth]{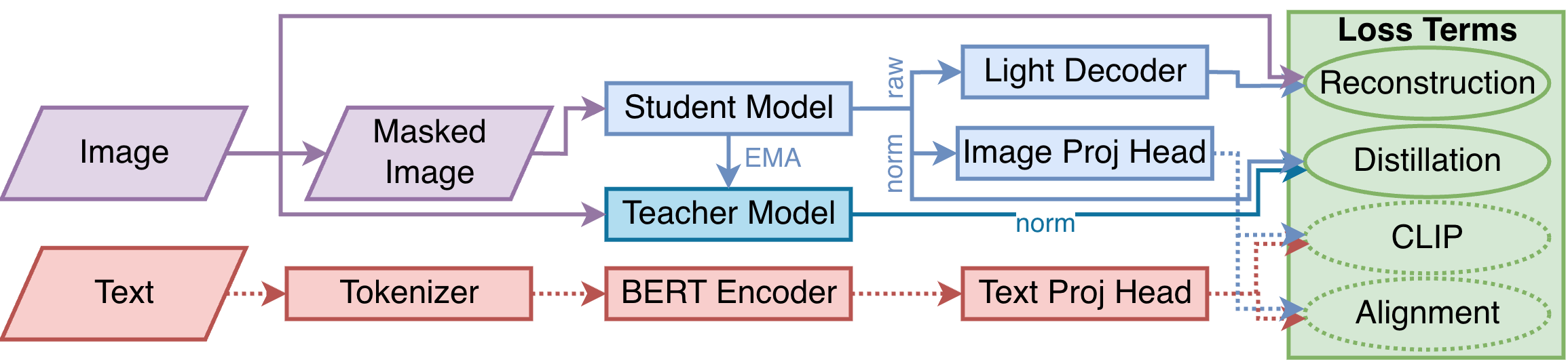}
       \caption{\textbf{Overview of pretraining framework.} A student-teacher architecture (UNet or SwinUNETR) is trained with masked image reconstruction, exponential moving average (EMA)-based distillation, where teacher weights are updated as a moving average of the student, and CLIP-style image-text alignment using BERT-encoded text embeddings \cite{ref_clip}. Dashed lines are associated with the text branch which can be removed for image-only model pretraining.}
    \label{fig:pretrain_finetune_overview}
\end{figure}

Prior 2D approaches fail to leverage volumetric context, often producing jagged artifacts along the axial dimension \cite{ref_cellpose,ref_stardist}. To address these limitations, we introduce a \textit{3D foundation model} for LSM that leverages unannotated images and free-text descriptions to learn rich multimodal representations. Our pretraining combining image modeling and reconstruction with image-text alignment allows the model to capture both structural and semantic information. The resulting pretrained backbone can be finetuned with minimal annotated data to perform segmentation, classification, and deblurring. An overview of the pretraining strategy is presented in Fig.~\ref{fig:pretrain_finetune_overview}. Our contributions are threefold:
\begin{enumerate}

    \item We curate a heterogeneous dataset of unannotated 3D LSM volumes across multiple organisms, stains, and imaging protocols, and compose corresponding textual descriptions for each sample.
    \item We utilize our assembled vision-language dataset to pretrain an LSM foundation model that can be finetuned for various downstream tasks.
    \item We evaluate image+text and image-only pretraining across downstream tasks and show that pretrained models outperform trained-from-scratch models and baselines. 
\end{enumerate}

\section{Method}

\subsection{Pretraining Data}
\textbf{Datasets} To promote generalization across organisms, structures, and imaging protocols, we assembled a heterogeneous pretraining dataset from multiple internal and publicly available resources.  

The internal dataset contains 24 single-channel mouse whole-brain images, each with a distinct immunostain. From each image, ten representative patches were manually selected. These training patches will be made public. 

\begin{figure} [t]
    \includegraphics[width=\textwidth]{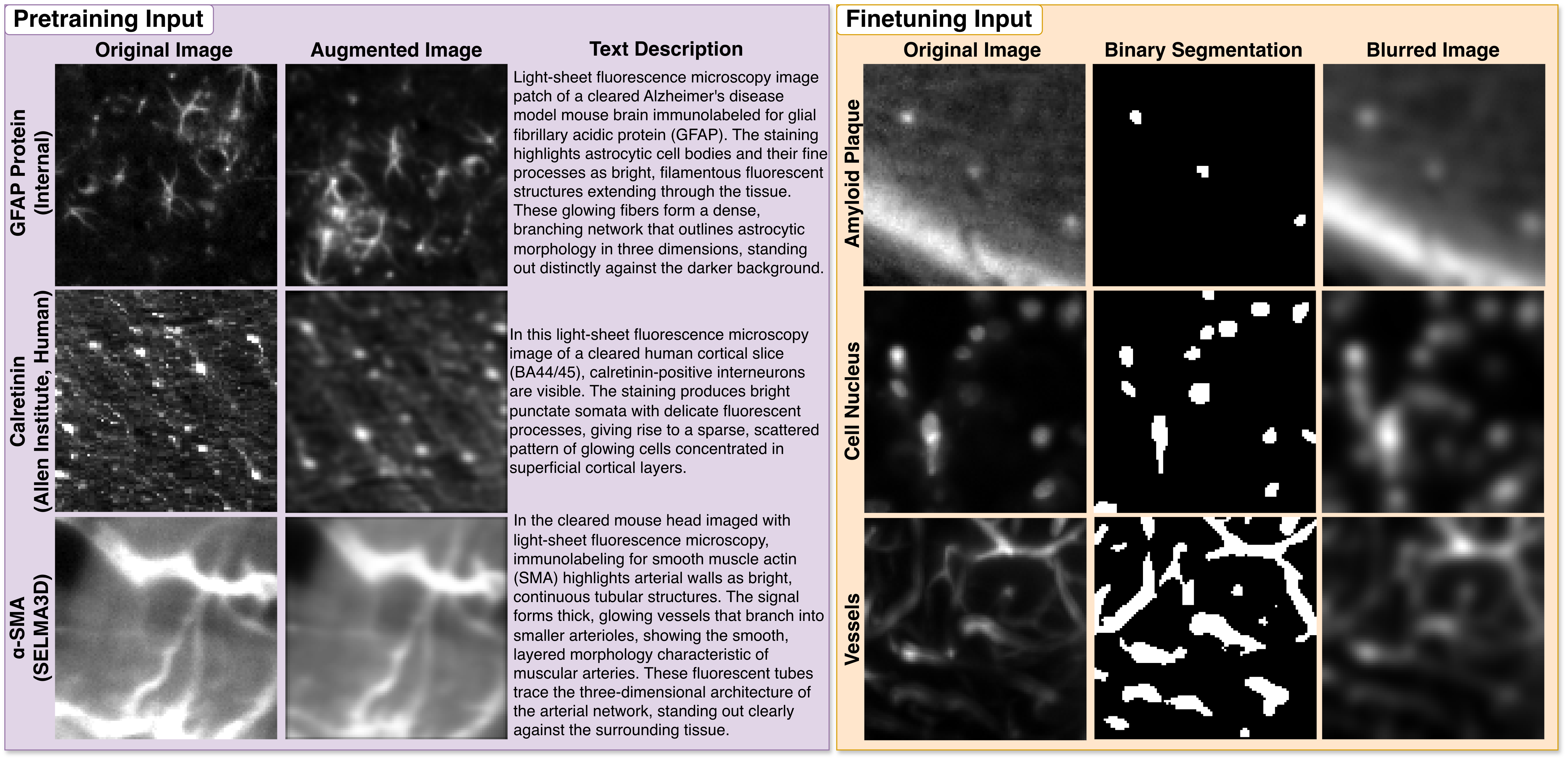}
    \caption{\textbf{Examples of input data} used during pretraining (left), and finetuning (right). During pretraining, each volumetric patch is paired with a text description, forming image-text pairs for contrastive learning. During finetuning, the pretrained encoder is adapted for downstream tasks using paired images and corresponding labels, including binary masks for segmentation and synthetically blurred images for deblurring.}
    \label{fig:data_overview}
\end{figure}

Public datasets include the SELMA3D MICCAI challenge dataset \cite{ref_selma3d2024,ref_selma3d2025} and three datasets from the Allen Institute \cite{ref_allen_dev_mouse,ref_allen_conn_proj,ref_allen_human2}. The SELMA3D dataset includes 30 single-channel mouse and human brain images of diverse structures, including neurons, cell nuclei, amyloid plaques, chondrocytes, and astrocytes, and 9 dual-channel vessel images. The datasets from the Allen Institute, a human brain dataset \cite{ref_allen_human2}, a developing mouse brain dataset \cite{ref_allen_dev_mouse}, and a viral labeling dataset of projection neurons \cite{ref_allen_conn_proj}, each include numerous multi-channel images.  From each public dataset, up to 10 non-overlapping patches were randomly selected per channel from each image, ensuring a minimum foreground threshold of 5\%. In total, 1,023 volumetric $96^3$ voxel patches were collected.\\
\textbf{Augmentations} All pretraining images were normalized using percentile-based intensity rescaling to map intensities to the [0, 1] range. To increase robustness and variability, extensive data augmentation was applied to each patch, including random flips, rotations, affine transformations, Gaussian noise, Gaussian smoothing, intensity scaling, and intensity shifting. Intensities were clamped back to [0, 1] after augmentation. Examples are shown in Fig.~\ref{fig:data_overview}.\\
\textbf{Text Prompts} For each volume, domain experts composed a structured 2-4 sentence free text caption describing biologically and visually salient features. Captions describe the staining target (e.g., protein or cell type), imaging modality and channel, specimen source (organism and tissue), morphological characteristics (e.g., filamentous, punctate, tubular), spatial organization within the tissue, and relevant pathological context. This process ensured that descriptions reflected both domain-specific terminology and observable image features. To increase linguistic diversity while preserving semantic fidelity, captions were paraphrased using an LLM \cite{ref_chatgpt}. Examples are shown in Fig.~\ref{fig:data_overview}.

\subsection{Finetuning Data}
\textbf{Datasets} For downstream segmentation and deblurring experiments, three annotated datasets from the SELMA3D challenge were used \cite{ref_selma3d2024,ref_selma3d2025}. Datatypes include amyloid plaque (single-channel), cell nucleus (single-channel), and vessel (dual-channel). Per structure and channel, up to 25 patches of size $96^3$ with sufficient foreground representation were randomly sampled, yielding a total of 84 annotated patches. Examples are shown in Fig.~\ref{fig:data_overview}. 

The finetuning datasets were selected to represent distinct structural regimes common in LSM, allowing us to evaluate the generality of the learned representations. Amyloid plaques and cell nuclei both correspond to isolated foreground objects but differ in foreground density and spatial distribution, with amyloid plaques being sparse, punctate structures as compared with the dense foreground distribution of cell nuclei. Vessel images were included to represent a complementary domain characterized by contiguous, elongated structures. Unlike isolated objects, vascular structures require modeling long-range spatial continuity and topology, posing distinct challenges \cite{ref_cldice}. 

To increase the number of stain classes for classification experiments, images from three additional datasets were included, in addition to those used for segmentation and deblurring: c-Fos positive active neurons from SELMA3D \cite{ref_selma3d2024}, mouse brain images from the mesoSPIM initiative \cite{ref_mesospim}, and human brain images from the Allen Institute \cite{ref_allen_human1}. In images with multiple channels, each stain was treated independently, and patches from all images were randomly selected ensuring sufficient foreground, resulting in 12 categories of images for classification.\\ 
\textbf{Augmentations} The intensity values in finetuning images were normalized identically to pretraining, and moderate augmentations including random flips, rotations, affine transformations, and Gaussian noise, were applied as well. 

\subsection{Model Architecture}

\textbf{Model Backbones} We develop our foundation models using two 3D architecture families: a convolutional encoder-decoder model, UNet \cite{ref_unet}, and a hierarchical transformer-based model, SwinUNETR \cite{ref_swinunetr,ref_monai}. We instantiate the backbone twice for each architecture in order to construct a student-teacher framework. During training, we provide the student with masked $96^3$ input volumes, where a randomized subset of non-overlapping patches of size $4^3$ or $8^3$ was zeroed out, while the teacher processes the corresponding unmasked volume. Rather than updating the teacher through gradient descent, we maintain its weights as an exponential moving average (EMA) of the student parameters. This asymmetric design encourages the student to infer stable representations at masked locations, while mitigating representation collapse \cite{ref_teacher_student,ref_dino}. \\
\textbf{Pretraining Objective} The pretraining objective combines four complementary loss terms. (1) A knowledge distillation loss computes the KL divergence between the student and teacher feature distributions at masked positions, encouraging the student to recover the teacher's full-context representations from partial observations \cite{ref_ibot}. (2) A lightweight convolutional decoder, supervised by an L1 reconstruction loss on masked regions, maps the student's raw feature map back to voxel space. (3) A cosine alignment loss minimizes the angular distance between image embeddings and text embeddings produced using a pretrained BERT-based text encoder \cite{ref_bert}. (4) A CLIP-style symmetric cross-entropy contrastive loss pushes matching image-text pairs together and non-matching pairs apart, across the batch \cite{ref_clip}. The framework supports image-only pretraining by disabling the text stream, whereby the alignment and contrastive losses are zeroed and only the distillation and reconstruction objectives remain active. 

The total loss is computed as a weighted sum of all four loss terms, with independently tunable coefficients for each component, as presented in Eq.~\eqref{eq:total_loss}. The combined gradient signal from the total loss is backpropagated through the student encoder, image projection head, light decoder, and (post-warmup) the unfrozen text encoder layers, while the teacher remains outside the computational graph. 

\begin{equation}
\label{eq:total_loss}
    \mathcal{L}_{\text{total}}
    =
    \lambda_{\text{dist}} \, \mathcal{L}_{\text{dist}}
    +
    \lambda_{\text{rec}} \, \mathcal{L}_{\text{rec}}
    +
    \lambda_{\text{align}} \, \mathcal{L}_{\text{align}}
    +
    \lambda_{\text{clip}} \, \mathcal{L}_{\text{clip}}
\end{equation}

where $\gamma_{\text{dist}}, \gamma_{\text{rec}}, \gamma_{\text{align}},$ and $\gamma_{\text{clip}}$ are independently tunable coefficients.\\

\noindent\textbf{Training} All trainable parameters are optimized jointly with AdamW. Training spanned 10-12 hours (280-500 epochs) on 2 A100/H100 GPUs using a 90/10 train-validation split. Loss weights and other hyperparameter values were selected via a sweep.  Details are available in our GitHub: \url{https://github.com/AdinaScheinfeld/lsm_fm_public_repo.git}.

\section{Finetuning and Experiments}

We consider three downstream tasks: Voxel-wise binary \textit{segmentation}, patch-level stain-type \textit{classification}, and \textit{deblurring} of degraded patches. For each task, both the UNet and the SwinUNETR pretrained backbones were evaluated.\\
\begin{table} [t]
\vspace{-0.2cm}
\centering
\fontsize{7}{7}\selectfont
\setlength{\tabcolsep}{2pt}
\renewcommand{\arraystretch}{1.05}
\caption{Segmentation performance in a few-shot (number of train patches=5) and many-shot (number of train patches=15) settings. Values are averaged across two held-out test patches from each of three CV folds. Overall, pretrained models outperform trained-from-scratch counterparts and strong baselines.}
\vspace{-2ex}
\label{tab:segmentation_results_fewmany}
\begin{tabular}{@{}>{\centering\arraybackslash}p{0.55cm} l cccccccccccc@{}}

\toprule
 & & \multicolumn{4}{c}{Amyloid Plaque} & \multicolumn{4}{c}{Cell Nucleus} & \multicolumn{4}{c}{Vessels} \\
 & & \multicolumn{2}{r}{Few-shot} & \multicolumn{2}{r}{Many-shot} & \multicolumn{2}{r}{Few-shot} & \multicolumn{2}{r}{Many-shot} & \multicolumn{2}{r}{Few-shot} & \multicolumn{2}{r}{Many-shot} \\
 & Method & Tot & Inst & Tot & Inst & Tot & Inst & Tot & Inst & Tot & Inst & Tot & Inst \\
\midrule

\grouplabel{5}{Bases} &
\makecell[l]{Cellpose (2D) \cite{ref_cellpose}} & -- & -- & -- & -- & 0.54 & 0.20 & 0.57 & 0.43 & -- & -- & -- & -- \\
& \makecell[l]{Cellpose (3D) \cite{ref_cellpose}} & -- & -- & -- & -- & 0.79 & 0.80 & \textbf{0.82} & 0.82 & -- & -- & -- & -- \\
& \makecell[l]{CellSeg3D \cite{ref_cellseg}} & -- & -- & -- & -- & 0.51 & 0.71 & 0.51 & 0.74 & -- & -- & -- & -- \\
& \makecell[l]{uSAM (b) \cite{ref_microsam}} & 0.66 & 0.41 & 0.71 & 0.47 & 0.49 & 0.00 & 0.52 & 0.10 & 0.86 & 0.77 & 0.90 & 0.81 \\
& \makecell[l]{uSAM (l) \cite{ref_microsam}} & 0.62 & 0.37 & 0.77 & 0.63 & 0.53 & 0.12 & 0.56 & 0.22 & 0.78 & 0.74 & 0.84 & 0.82 \\
\midrule

\grouplabel{2}{Scr} &
\makecell[l]{Swin} & 0.50 & 0.09 & 0.67 & 0.45 & 0.78 & 0.95 & 0.81 & 0.97 & 0.86 & 0.79 & 0.88 & 0.76 \\
& \makecell[l]{UNet} & 0.50 & 0.11 & 0.72 & 0.58 & 0.75 & 0.86 & 0.80 & 0.88 & 0.83 & \textbf{0.86} & 0.89 & 0.68 \\
\midrule

\grouplabel{2}{Img} &
\makecell[l]{Swin} & 0.58 & 0.21 & 0.79 & 0.73 & 0.76 & 0.94 & 0.80 & 0.94 & 0.87 & 0.82 & 0.89 & 0.81 \\
& \makecell[l]{UNet} & 0.50 & 0.03 & 0.50 & 0.00 & \textbf{0.80} & 0.87 & \textbf{0.82} & 0.96 & 0.61 & 0.56 & 0.89 & 0.84 \\
\midrule

\grouplabel{3}{Img+T} &
\makecell[l]{Swin} & 0.48 & 0.01 & 0.76 & 0.72 & \textbf{0.80} & 0.93 & 0.80 & \textbf{0.99} & 0.87 & 0.75 & \textbf{0.92} & 0.87 \\
& \makecell[l]{Swin (over)} & 0.61 & 0.47 & 0.79 & \textbf{0.83} & 0.79 & \textbf{0.96} & 0.81 & 0.98 & \textbf{0.89} & 0.84 & \textbf{0.92} & \textbf{0.88} \\
& \makecell[l]{UNet} & \textbf{0.68} & \textbf{0.58} & \textbf{0.80} & 0.69 & 0.78 & 0.91 & 0.81 & 0.94 & 0.86 & 0.76 & \textbf{0.92} & 0.81 \\

\bottomrule
\end{tabular}
\\[1.0ex]

\begin{minipage}{\linewidth}
{\footnotesize \textbf{Note}: \textbf{Tot}: total Dice; \textbf{Inst}: instance Dice; \textbf{Bases}: baselines; \textbf{Scr}: trained from scratch; \textbf{Img}: image-only pretraining; \textbf{Img+T}: image+text pretraining; \textbf{over}: overfit variant. Bold indicates best per column.}
\end{minipage}

\vspace{-0.3cm}
\end{table}\\
\textbf{Segmentation} For the UNet backbone, the pretrained encoder weights are mapped directly into a MONAI UNet. The encoder is frozen for an initial warmup period and subsequently finetuned at a reduced learning rate, while the final decoder block is trained at the full learning rate throughout. We instantiate a full SwinUNETR model, loading the pretrained encoder weights and freezing them during an initial warmup phase. The encoder is then unfrozen and finetuned at a lower learning rate than the convolutional decoder. UNet models were trained using Dice-Focal loss, while SwinUNETR models used Dice-Cross Entropy. Loss functions were chosen based on preliminary experiments to ensure stable optimization across architectures. For both, optimization was performed using AdamW, with early stopping based on validation Dice.\\
\textbf{Classification} The pretrained encoder is finetuned with a linear classification head appended to the bottleneck representation. The deepest encoder feature map is globally average-pooled to produce a fixed-size descriptor, which is then projected to class logits via a single linear layer. The encoder is frozen for an initial warmup period, then optimized at a reduced learning rate, while the classification head is trained at the full learning rate throughout. Cross-entropy loss with inverse-frequency class weighting is applied to account for class imbalance. Models were optimized using AdamW with early stopping (validation accuracy).\\
\textbf{Deblurring} For each datatype, we provide paired sharp and synthetically\break blurred image patches. The model is trained to reconstruct the sharp volume from the blurred input by predicting an additive residual that is summed with the blurred image and clamped to the range [0,1]. The training objective combines four terms: an L1 reconstruction loss, a structural similarity loss (SSIM), a 3D gradient-based edge consistency loss that penalizes discrepancies in spatial gradient magnitudes, and a high-frequency loss term to preserve fine structural details. Both backbones are instantiated as full encoder-decoder networks and initialized from the respective pretrained checkpoints. Optimization is performed using AdamW with early stopping based on validation total loss.\\
\textbf{Experimental Design} To analyze scaling behavior, we evaluated finetuning via 3-fold cross-validation across few-shot and many-shot regimes. Training sizes ranged from 5-15 patches per datatype for segmentation and deblurring, and 56-105 samples for classification. Across all tasks, data was randomly split 80/20 for training and validation, with 2 patches per datatype held out for testing.
\noindent\textbf{Baselines} Segmentation baselines include other pretrained models, $\mu$SAM (base and large) \cite{ref_microsam} and CellSeg3D \cite{ref_cellseg}, and a state-of-the-art fully supervised model, Cellpose-SAM (2D and 3D variants) \cite{ref_cellpose}. Classification baselines include a non-deep-learning approach, PCA, and a standard deep-learning-based model, 3D ResNet-18 \cite{ref_resnet}. Additionally, a model of each backbone type was trained from scratch for each task as a directly comparable non-pretrained comparison.\\
\textbf{Overtraining} To assess whether pretraining epochs limited downstream performance, we conducted an overtraining experiment where the SwinUNETR model was trained beyond validation convergence ($\sim$4,000 epochs), effectively overfitting \cite{ref_overtrain}. The final checkpoint from this model was then finetuned as before. 

\begin{table} [t]
\vspace{-0.2cm}
\centering
\fontsize{7}{7}\selectfont
\setlength{\tabcolsep}{2pt}
\caption{Classification and deblurring performance in few-shot and many-shot settings. Few/many correspond to train patches=56/105 for classification and train patches=5/15 for deblurring. Values are averaged across held-out test patches from three CV folds. For deblurring, PSNR (not shown) exhibits similar trends as SSIM.}
\vspace{-2ex}
\label{tab:classification_and_deblur_results} 

\begin{tabular}{@{}>{\centering\arraybackslash}p{0.55cm} l 
cccc ||
cccccc@{}}
\toprule

& & \multicolumn{4}{c||}{Classification} 
& \multicolumn{6}{c}{Deblurring (SSIM)} \\

& & \multicolumn{2}{c}{Few} 
& \multicolumn{2}{c||}{Many}
& \multicolumn{2}{c}{Amyloid Plaque} 
& \multicolumn{2}{c}{Cell Nucleus}
& \multicolumn{2}{c}{Vessels} \\

& Method 
& ACC & MACRO $F_1$ 
& ACC & MACRO $F_1$
& Few & Many
& Few & Many
& Few & Many \\

\midrule

\grouplabel{2}{Bases} &
\makecell[l]{PCA}\rule{0pt}{2.8ex} & 0.33 & 0.28 & 0.47 & 0.40 & -- & -- & -- & -- & -- & -- \\
& \makecell[l]{ResNet\cite{ref_resnet}}\rule{0pt}{2.8ex} & 0.36 & 0.27 & 0.49 & 0.43 & -- & -- & -- & -- & -- & -- \\

\midrule

\grouplabel{2}{Scr} &
\makecell[l]{Swin} 
& 0.40 & 0.33 & 0.47 & 0.44
& 0.61 & 0.66
& 0.79 & 0.85
& 0.80 & 0.90 \\

& \makecell[l]{UNet} 
& 0.49 & 0.46 & 0.61 & 0.57
& 0.63 & 0.67
& 0.81 & 0.87
& 0.88 & 0.90 \\

\midrule

\grouplabel{2}{Img} &
\makecell[l]{Swin} 
& 0.51 & 0.46 & 0.53 & 0.48
& 0.57 & 0.65
& 0.81 & 0.86
& 0.73 & 0.69 \\

& \makecell[l]{UNet} 
& \textbf{0.71} & \textbf{0.69} & 0.62 & 0.58
& 0.64 & \textbf{0.69}
& \textbf{0.86} & \textbf{0.89}
& \textbf{0.89} & \textbf{0.92} \\

\midrule

\grouplabel{2}{Img+T} &
\makecell[l]{Swin}\rule{0pt}{3.0ex}
& 0.40 & 0.35 & 0.65 & 0.61
& 0.58 & 0.64
& 0.79 & 0.85
& 0.85 & 0.90 \\

& \makecell[l]{UNet}\rule{0pt}{3.0ex}
& 0.61 & 0.57 & \textbf{0.74} & \textbf{0.69}
& \textbf{0.65} & \textbf{0.69}
& 0.80 & \textbf{0.89}
& 0.78 & 0.90 \\

\bottomrule
\end{tabular}

\vspace{1ex}

\begin{minipage}{\linewidth}
{\footnotesize \textbf{Note}: \textbf{Bases}: baselines; \textbf{Scr}: trained from scratch; \textbf{Img}: image-only pretraining; \textbf{Img+T}: image+text pretraining; \textbf{ACC}: accuracy. Bold indicates best per column.}
\end{minipage}

\vspace{-3ex}
\end{table}

\begin{figure} [t]
    \centering
    \includegraphics[width=0.9\textwidth]{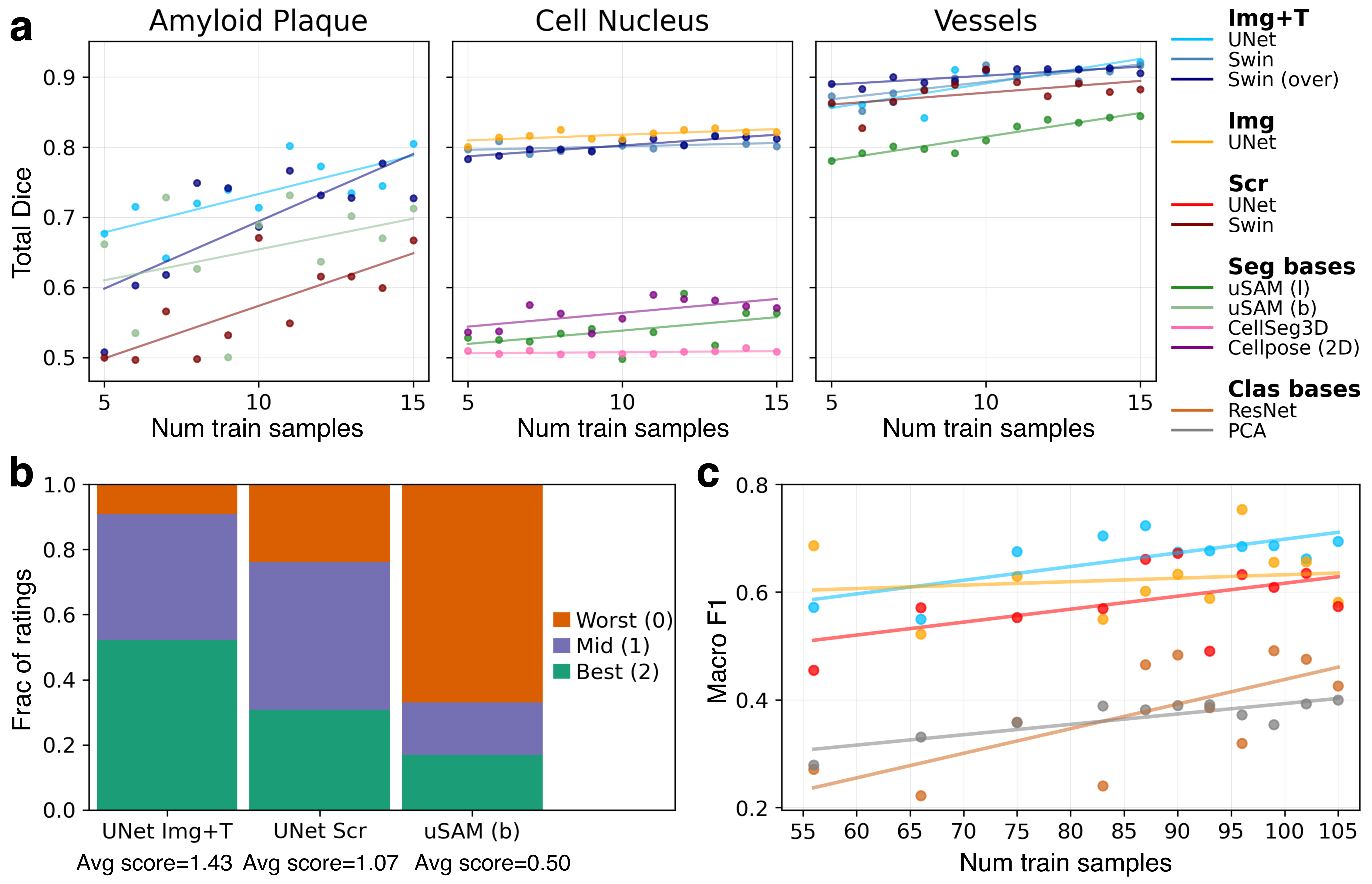}
    \caption{\textbf{Inference results} \textbf{(a)} Segmentation performance (total Dice) across train sizes. Points show cross-validation averages on held-out patches. Pretrained models outperform scratch-trained models and baselines across datatypes. \textbf{(b)} Blinded expert evaluation of 60 segmentation samples. Bars show the distribution of best/middle/worst rankings (2/1/0 points), with average scores reported below each. The pretrained model achieves highest overall and average scores. \textbf{(c)} Classification performance (macro $F_1$) across train sizes, averaged over cross-validation splits. Points denote mean performance on held-out patches. Pretrained models outperform scratch-trained models and baselines. \textbf{Note}: \textbf{Img+T}: image+text pretraining; \textbf{Img}: image-only pretraining; \textbf{Scr}: trained from scratch; \textbf{uSAM (l/b)}: large/base uSAM models; \textbf{over}: overfit variant.}
    \label{fig:results}
\end{figure}

\section{Results}

\textbf{Metrics-Based Evaluation} Across all tasks and datasets, pretrained models consistently outperform trained-from-scratch counterparts and task-specific baselines, particularly in low-data regimes. Image-only pretraining already yields substantial gains, while image-text pretraining provides additional improvements in several settings, especially for segmentation.

Segmentation results from pretrained models show marked improvements in total Dice and instance Dice across amyloid plaque, nucleus, and vessel datasets as compared with trained-from-scratch models and baselines. Few-shot and many-shot results on held-out test sets are summarized in Table~\ref{tab:segmentation_results_fewmany} and extended results are presented in Fig.~\ref{fig:results}. Similarly, classification performance improves in both accuracy and macro~F1, with complementary strengths between image+text and image-only models, depending on train set size. Few-shot and many-shot results on held-out test sets are summarized in Table~\ref{tab:classification_and_deblur_results} and extended results are presented in Fig.~\ref{fig:results}. For deblurring, pretrained models achieve higher SSIM and PSNR scores, indicating improved perceptual quality and fidelity. Results on held-out test sets are reported in Table~\ref{tab:classification_and_deblur_results}.

 Furthermore, across most downstream tasks, performance differences between the overtrained and normally trained backbones were modest, indicating that standard pretraining was sufficient. However, in select experiments, the overtrained backbone yielded small but consistent improvements.\\
 \noindent\textbf{Human Expert Evaluation} Qualitative evaluation by 6 domain experts, each with 5-10 years of experience with LSM images, was conducted as an external validation of the quantitative findings. In blinded assessments, each expert independently evaluated 60 segmentation predictions spanning multiple datatypes. Across experts and datatypes, the pretrained model was most frequently selected as the best-performing approach, reinforcing the improvements observed in the metrics-based analysis and demonstrating that the quantitative gains correspond to perceptible improvements in segmentation quality, as highlighted in Fig.~\ref{fig:results}.
 
\section{Discussion and Conclusion}

We present a foundation model for light sheet microscopy that leverages unannotated volumetric data and text descriptions to enable efficient and improved few-shot and many-shot learning across downstream tasks. Our findings demonstrate that foundation model pretraining enables effective transfer to segmentation, classification, and deblurring tasks with minimal annotation. By reducing reliance on extensive annotations and improving generalization across stains and structures, our approach offers a practical pathway toward scalable and reusable analysis pipelines for large-scale microscopy data. 

The strong performance of image-only pretrained models suggests that structural cues alone are highly informative, while text used in image+text models provides complementary semantic guidance that further improves results in specific scenarios. Moreover, while excessive pretraining is not strictly necessary, extended exposure to diverse unannotated volumes may further refine representations in certain settings, without degrading generalization. 

\vspace{0.2cm}
\noindent\textbf{Acknowledgement} Research data generation from Zhuhao Wu laboratory in this report was supported by the NIH grant R01AI158676, R01MH131537, \break RF1MH128969 and R01MH142410.\\
\noindent\textbf{Disclosure of Interests} The authors have no competing interests to declare that are relevant to the content of this article.

%
%
%
%

\end{document}